\documentclass[letterpaper]{article} 

\PassOptionsToPackage{table}{xcolor}

\usepackage{aaai2026}  
\usepackage{times}  
\usepackage{helvet}  
\usepackage{courier}  
\usepackage[hyphens]{url}  
\usepackage{graphicx} 
\usepackage{natbib}  
\usepackage{caption} 
\usepackage{algorithm}
\usepackage{algorithmic}

\usepackage{listings}
\usepackage{xcolor} 
\usepackage{multirow}
\usepackage{amsmath} 
\usepackage{booktabs} 
\usepackage{amssymb}

\newcommand{\method}{LaMPE} %

\usepackage{newfloat}
\usepackage{listings}
\DeclareCaptionStyle{ruled}{labelfont=normalfont,labelsep=colon,strut=off} 
\floatstyle{ruled}
\newfloat{listing}{tb}{lst}{}
\floatname{listing}{Listing}

\title{\method{}: Length-aware  Multi-grained Positional Encoding for Adaptive  Long-context Scaling Without Training}
\author{
    Sikui Zhang\textsuperscript{\rm 1 2 3},
    Guangze Gao\textsuperscript{\rm 1 2},
    Ziyun Gan\textsuperscript{\rm 1 2},
    Chunfeng Yuan\textsuperscript{\rm 1 2}\thanks{Corresponding author.},\\
    Zefeng Lin\textsuperscript{\rm 1 2},
    Houwen Peng\textsuperscript{\rm 4},
    Bing Li\textsuperscript{\rm 1 2},
    Weiming Hu\textsuperscript{\rm 1 2}
}
\affiliations{
    \textsuperscript{\rm 1}Beijing Key Laboratory of Super Intelligent Security of Multi-Modal Information, CASIA \\
    \textsuperscript{\rm 2}State Key Laboratory of Multimodal Artifcial Intelligence Systems, CASIA \\
    \textsuperscript{\rm 3}School of Artificial Intelligence, University of Chinese Academy of Sciences \\
     \textsuperscript{\rm 4}Tencent  \\
    \{zhangsikui2024, gaoguangze2023, ganziyun2025, zefeng.lin\}@ia.ac.cn\\
    \{cfyuan, bli, wmhu\}@nlpr.ia.ac.cn, penghouwen@icloud.com
}

\copyrighttext{Preprint. Under review.}

\usepackage{bibentry}

\begin{document}

\maketitle

\definecolor{myred}{RGB}{255,171,163}
\definecolor{myblue}{RGB}{86,193,255}

\begin{abstract}
Large language models (LLMs) experience significant performance degradation when the input exceeds the pretraining context window, primarily due to the out-of-distribution (OOD) behavior of Rotary Position Embedding (RoPE). Recent studies mitigate this problem  by remapping OOD positions into the in-distribution range with fixed mapping strategies, ignoring the dynamic relationship between input length and the model’s effective context window. To this end, we propose \textbf{L}ength-\textbf{a}ware \textbf{M}ulti-grained \textbf{P}ositional \textbf{E}ncoding (\method{}), a training-free method that fully utilizes the model’s effective context window for adaptive long-context scaling in LLMs. Motivated by the left-skewed frequency distribution of relative positions, \method{} establishes a dynamic relationship between mapping length and input length through a parametric scaled sigmoid function to adaptively allocate positional capacity across varying input lengths. Meanwhile, \method{} devises a novel multi-grained attention mechanism that strategically allocates positional resolution across different sequence regions to capture both fine-grained locality and long-range dependencies. Our method can be seamlessly applied to a wide range of RoPE-based LLMs without training. Extensive experiments on three representative LLMs across five mainstream long-context benchmarks demonstrate that \method{} achieves significant performance improvements compared to existing length extrapolation methods.  The code will  be released at \texttt{https://github.com/scar-on/LaMPE}.
\end{abstract}

\section{Introduction}
Rotary Position Embedding (RoPE) has become a cornerstone technique in large language models (LLMs) due to its excellent performance~\cite{su2024roformer,touvron2023llama2,yang2024qwen2technicalreport,dubey2024llama}. However, RoPE-based LLMs  are typically limited  by the context window formed during pretraining and post-training stages. When the input length exceeds the context window, RoPE encounters out-of-distribution (OOD) relative positions, as larger relative positions were not seen during the training phase~\cite{han2024lminfinit,kexuefm-9444}. Consequently, this often leads to significant degradation in model performance, limiting the model's applicability in real-world tasks, such as repository-level code completion~\cite{guo2024deepseekcoder}, long-document question-answering~\cite{li2024longcontextllmsstrugglelong}, and summarization~\cite{pratapa-mitamura-2025-scaling}.

\begin{figure}[!tb]
    \centering
    \includegraphics[width=0.95\linewidth]{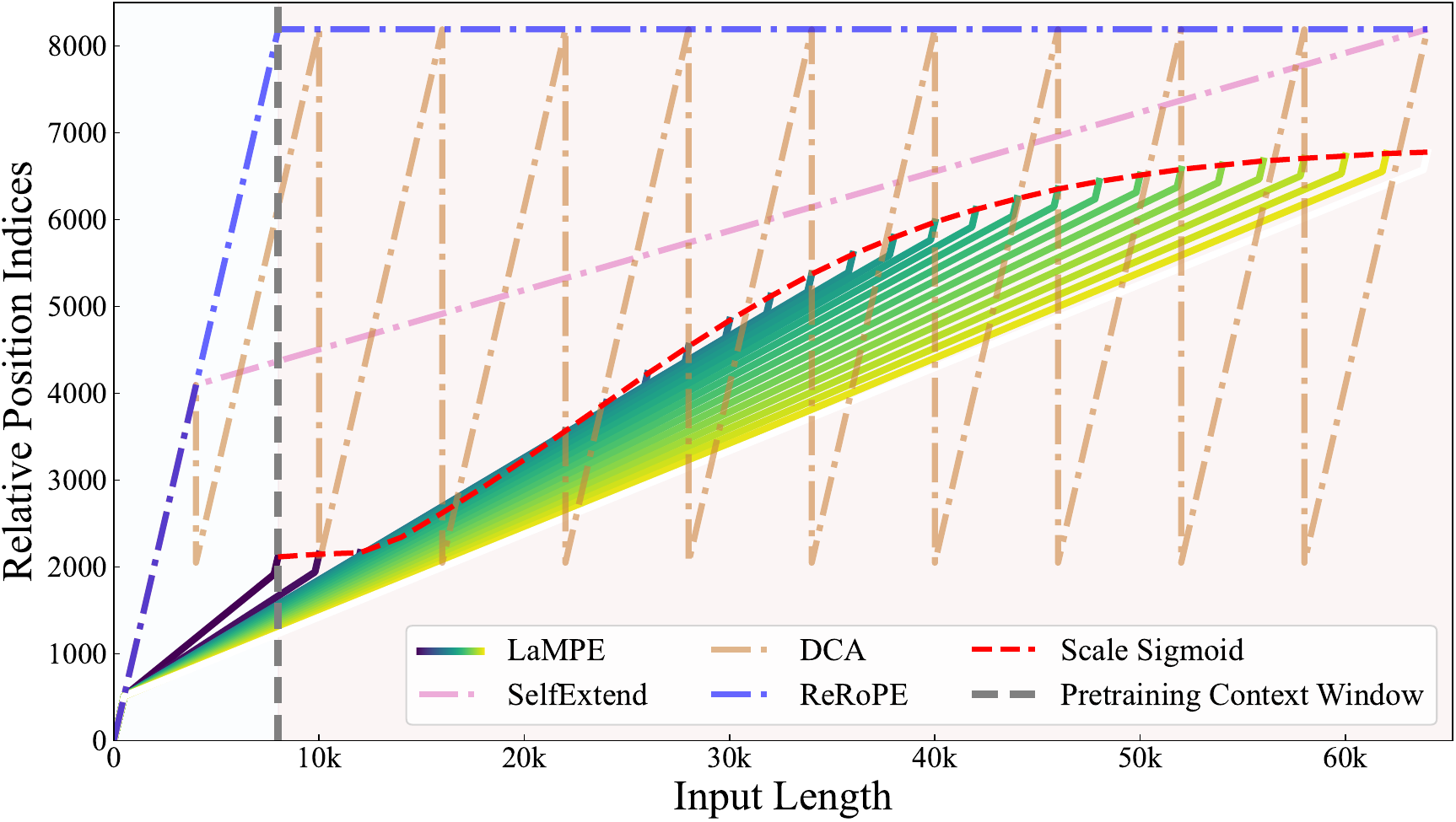}
    \caption{Comparison of different mapping strategies generated by \method{} and other extrapolation methods, including ReRoPE, SelfExtend, and DCA.}
    \label{fig:method-eg}
\end{figure}

Existing  RoPE-based methods for LLM  extrapolation can be broadly categorized into two types: base-modified and position indices modified. The first type focuses on modifying the base~\cite{NTK,peng2023yarnefficientcontextwindow,pi,codellama} to extend the context window. However, these methods typically require additional fine-tuning on longer texts and have explicit extrapolation upper bounds. In contrast, the second type that we focus on in this work achieves training-free length extrapolation, including ReRoPE~\cite{kexuefm-9708}, SelfExtend~\cite{pmlr-v235-jin24b}, and DCA~\cite{an2024training}. The core idea is to remap OOD relative positions into the pretraining context window with fixed mapping strategies. 

\begin{figure*}[!tb]
    \centering
    \includegraphics[width=0.95\linewidth]{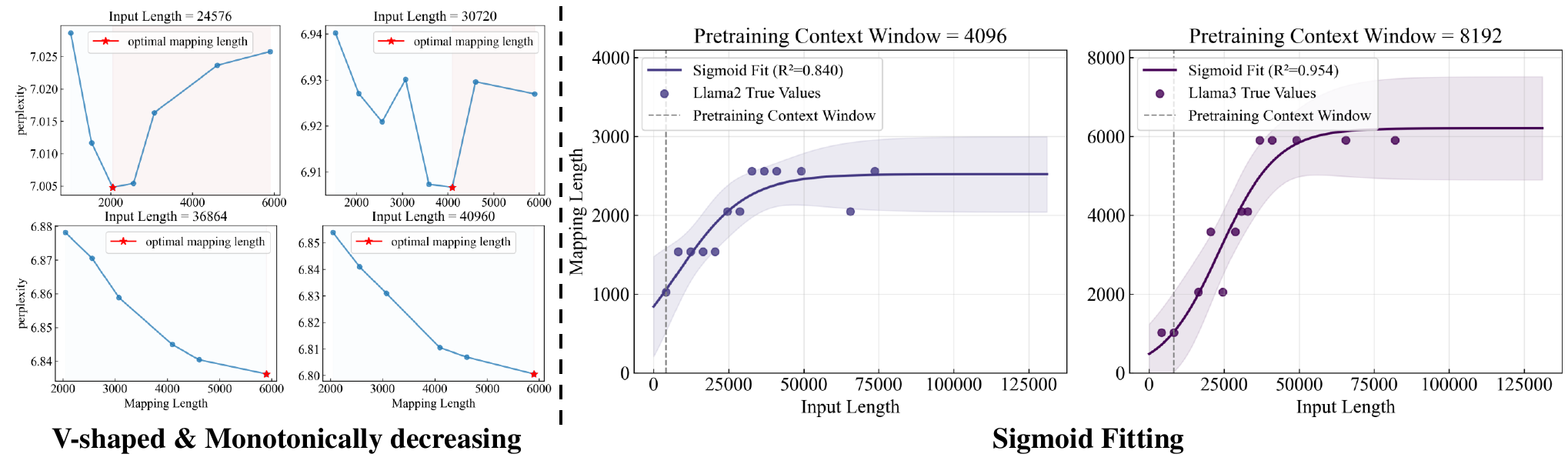}
    \caption{\textbf{Left:} Visualization of the variation in perplexity with respect to mapping length. \textbf{Right:} Sigmoid fit of the relationship between input length and optimal mapping length for Llama2 and Llama3.}
    \label{fig:fit}
\end{figure*}

As shown in Figure~\ref{fig:method-eg}, although these studies have shown promising results in mitigating the OOD issue compared to the first type of methods, they still face two major challenges corresponding to the peak point and slope of the curve respectively. (1) Overlooking Skewed  Training Distribution: Such fixed mappings inherently overlook the left-skewed frequency distribution of relative positions~\cite{an2024does} formed during the training phase. Specifically, they treat all positions within the context window as equally well-trained, while in reality, long-range positions are severely undertrained compared to short-range ones. (2) Inflexible Static Grouping: Fixed group sizes  are used for position mapping, regardless of varying input lengths. This lack of adaptability prevents the optimal utilization of well-trained short-range positions, as the same grouping granularity cannot be optimal for both short and long sequences. These limitations highlight the need for a more flexible strategy that can adaptively set the optimal mapping length according to the input length. Thus, a key question arises: \textit{How can we adaptively set the optimal mapping length and group size  based on  input length?}

 To this end, we propose \textbf{L}ength-\textbf{a}ware  \textbf{M}ulti-grained \textbf{P}ositional \textbf{E}ncoding (\method{}), a training-free method that fully utilizes the model's effective context window for length extrapolation.  \method{} employs a parametric scaled sigmoid function to adjust the effective position mapping, enabling adaptive allocation of positional capacity across varying input lengths. This design is motivated by our empirical observation that the relationship between input length and optimal mapping length follows an S-shaped trend, as shown in Figure~\ref{fig:fit}(Right). Moreover, drawing inspiration from the critical roles of neighboring tokens and initial tokens~\cite{peysakhovich2023attentionsorting, liu2023lost, xiao2024efficientstreaming}, we propose a multi-grained mechanism to ensure that current tokens maintain fine-grained positional information with both neighboring tokens and initial tokens. Owing to its training-free and plug-and-play nature, \method{} can  be seamlessly applied to any RoPE-based LLM.

We comprehensively evaluate  \method{} on a wide range of long-context benchmarks, including LongBench~\cite{bai2024longbench}, L-Eval~\cite{L-eval}, $\infty$Bench~\cite{infitebench}, RULER~\cite{hsieh2024ruler} and PG-19~\cite{pg19}. Our experiments cover LLMs with context windows ranging from 4K to 128K, including Llama2-7B-Chat~\cite{touvron2023llama2}, Llama3-8B-Instruct~\cite{meta2024introducing}, and Llama3.1-8B-Instruct~\cite{dubey2024llama}. The experimental  results demonstrate that \method{} consistently outperforms existing extrapolation methods both at extrapolated lengths and  within the pretraining context window. Our main contributions are summarized as follows:
\begin{itemize}
    \item We propose \method{}, a training-free method that incorporates a dynamic  mapping strategy and multi-grained attention mechanism for adaptive long-context scaling.
    \item We explore perplexity variations across different mapping lengths, identifying V-shaped and monotonically decreasing patterns that guide position remapping.
    \item Comprehensive evaluation demonstrates that \method{} outperforms existing extrapolation methods on both real-world and synthetic long-context tasks.
\end{itemize}

\section{Background}
\label{Sec:Background}

\subsection{Rotary Position Embedding (RoPE)}
RoPE was first proposed in RoFormer~\cite{su2024roformer}, where it incorporates relative  positional information by applying a phase rotation to the Query and Key in attention computation. For the token embedding at position $i$, denoted as $x_i$, we use $f$ to represent the operation that injects positional information into $x_i$:
\begin{equation}
    f(x_i,i,\theta)=R^{\theta}_{i}W^Tx_i ,
\end{equation}
where $W$ denotes the learnable projection matrix of a linear layer, $R^{\theta}_{i}$ represents the rotary positional encoding matrix with the Ptolemy’s identity $R^{\theta}_{j-i} = (R^{\theta}_{i})^T R^{\theta}_{j}$, where $\theta$ represents frequency basis. So we define the dot product between the Query and Key at positions $i$ and $j$ as:
\begin{align}
    q^T_ik_j  & = f_{q}(x_i,i,\theta)^Tf_{k}(x_j,j,\theta) ,\\
    &= (R^{\theta}_{i}W^T_qx_i)^T (R^{\theta}_{j}W^T_k x_j) ,\\
    &= (W^T_qx_i)^T R^{\theta}_{j-i} (W^T_k x_j). 
\end{align}

During the attention computation, the positional information of the relative position $j-i$ can be incorporated by assigning the positional information of $i$ and $j$ to the Query and Key, respectively.

\begin{figure*}[htb]
    \centering
    \includegraphics[width=0.9\linewidth]{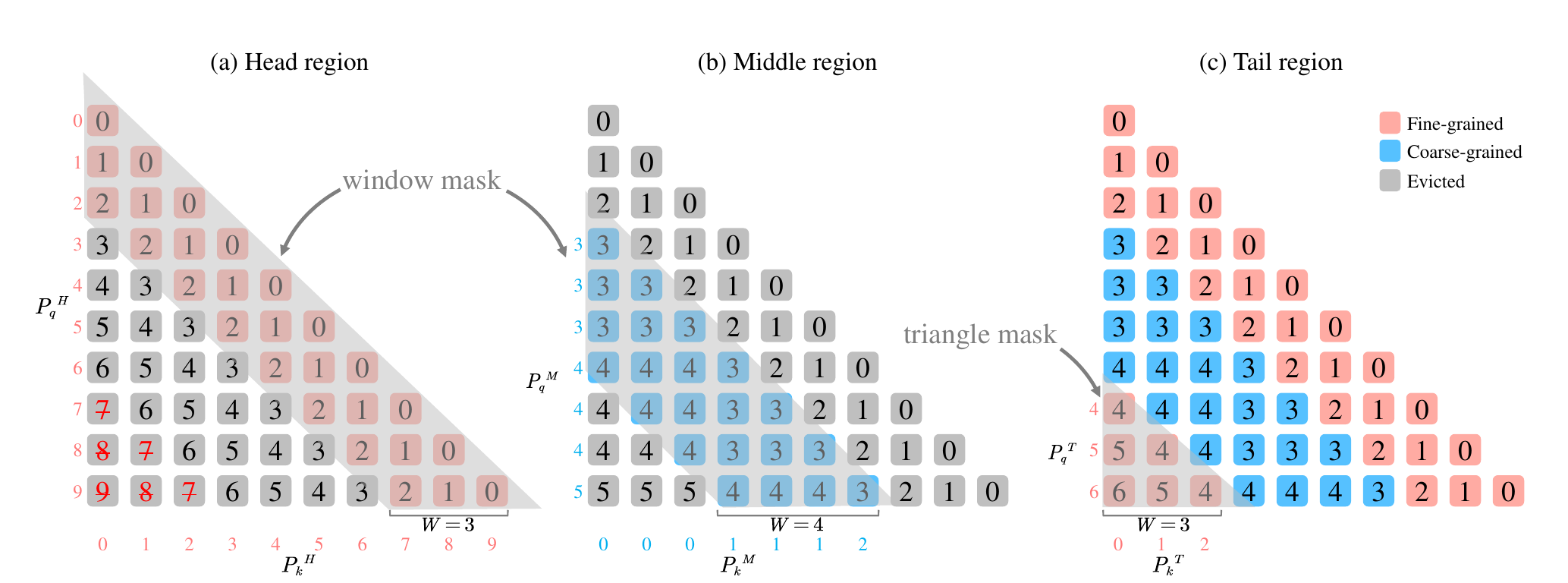}
    \caption{Illustration of \method{}'s Relative Position Matrix $PE$ for a sequence of length $l=10$ and pretraining context window $n = 7$. (a) head region with window size $s_1=3$. (b) middle region maps out-of-distribution relative positions ($[7, 8, 9]$) into the pretraining context window via the linear normalization. (c) tail region replaces the lower-left triangle part of the linear normalization with a window size of $s_2=3$.}
    \label{fig:method}
\end{figure*}

\subsection{Extrapolation of RoPE}
Previous work has focused on extending the context window by adjusting the rotary matrix $\theta$ in the function $f(x_i,i,\theta)$, including NTK-RoPE~\cite{NTK}  and YaRN~\cite{peng2023yarnefficientcontextwindow}. In contrast, another line of research focuses on adjusting  position indices $i$  to remap out-of-distribution relative positions  into the in-distribution range. For example, SelfExtend, for each token, selects the $w$ nearest tokens within a local window. Tokens outside this window are partitioned into multiple groups, each containing $G$ tokens, where all tokens in a group share the same position index. 
Given a fixed group size $G$ and local window size $w$, and a pretraining context window of size $n$, the model's context window is extended to $(n - w) \times G + w$.

\section{Method}
In this section, we present the details of Length-aware  Multi-grained Positional Encoding (\method{}). \method{} comprises two key components: (1) a Length-aware Mapping Strategy that establishes the relationship between input length and optimal mapping length through a parametric scaled sigmoid function, and (2) a Multi-grained Attention Mechanism  that partitions the input sequence into three distinct regions, each with tailored positional encoding granularity to jointly capture both locality and long-range dependencies.

\subsection{Length-aware Dynamic Mapping Strategy}
Existing methods typically adopt fixed mapping strategies that fail to generalize across inputs of varying lengths. To enable dynamic mapping, we first investigate the relationship between input length and optimal mapping length through empirical analysis.

\paragraph{Details.} We adopt the widely-used perplexity (PPL) as our evaluation metric and conduct experiments on the PG-19 dataset~\cite{pg19} using Llama2-7B-Chat and Llama3-8B-Instruct with pretraining context windows of 4K and 8K tokens respectively, where K denotes 1024. We systematically vary input length and mapping length to examine how perplexity changes under different configurations. Specifically, the input length is progressively increased from 4K to 64K. For each input length, the mapping length is varied from small values up to the pretraining context window.

\paragraph{Observations.} Our analysis reveals two key observations about how perplexity varies with mapping length. As shown in Figure~\ref{fig:fit}(Left): (1) \textbf{V-shaped} pattern for short inputs: When the input length is small, perplexity exhibits a V-shaped curve as mapping length increases. Starting from relatively high perplexity at small mapping lengths (e.g., 1/8 of the pretraining context window), perplexity gradually decreases to reach a minimum before rising again. This indicates a trade-off where both overly small and large mapping lengths lead to suboptimal performance. (2) \textbf{Monotonically decreasing} for long inputs: Once the input length exceeds a certain threshold, the V-shaped pattern disappears and perplexity consistently decreases with increasing mapping length, up to the pretraining context window size. These observations indicate that an optimal mapping length exists and varies depending on the input length. More observation results can  be found in  Appendix A.

\paragraph{Dynamic Mapping.} Based on the above observations, we devise a length-aware dynamic mapping strategy to adaptively determine the optimal mapping length using a scaled sigmoid function. As shown in Figure~\ref{fig:fit}(Right), the optimal mapping length exhibits an S-shaped growth pattern that can be effectively modeled by a scaled sigmoid function. This scaled sigmoid function can be expressed as follows:
\begin{equation}
    m=\frac{L}{1+e^{-(al+b)}},
    \label{eq:sigmoid}
\end{equation}
where $L$ denotes the maximum value of the curve, $l$ represents the input length, and $m$ corresponds to the optimal mapping length.  This sigmoid function provides a smooth and continuous approximation of the optimal mapping length across a wide range of input lengths.  The parameters $a$ and $b$ can be estimated via simple curve fitting based on a small number of  points, making this method highly practical for real-world applications.

\subsection{Multi-grained Attention Mechanism}
To address the limitations of fixed projected grouping strategies, we propose a multi-grained attention mechanism that strategically allocates positional resolution across different sequence regions based on the obtained optimal mapping length from the previous section. As illustrated in Figure~\ref{fig:method}, our method partitions the input sequence into three distinct regions, each with tailored positional encoding granularity. The following sections detail the adaptive grouping  and its integration with attention computation. 

\paragraph{Adaptive Grouping.}
Unlike fixed grouping methods,  our method dynamically adapts the relative position matrix \(PE \in {\mathbb{Z}}^{l\times l}\) based on input length, where each entry \(PE[i][j]\) remaps  the original relative position \((i-j)\) to a new relative position. Given input sequence length \(l\) and optimal mapping length \(m\), the updated relative position matrix is defined as:
\begin{equation}
\label{eq:ada_group}
    PE[i][j]=\left\lfloor \dfrac{m}{l}(i-j) \right\rfloor,\\
\end{equation}
where $\lfloor \cdot \rfloor$ denotes the floor operation. As shown in Figure~\ref{fig:method}(b), our method dynamically adapts the relative position matrix based on the input length, where the compression ratio ($\frac{m}{l}$) determines how original relative positions are remapped. This linear transformation effectively maps the original position range  $[0, l-1]$  to a compressed range $[0, m-1]$, enabling adaptive granularity control based on sequence length.

\paragraph{Restoring Locality and Long-Range Dependencies.}
In LLMs, key instructions or questions often appear at the beginning or end of the prompt. However,  only applying Eq.~\ref{eq:ada_group} may disrupt the model’s ability to capture local relationships and long-range dependencies  because it alters the relative positions between neighboring  and initial tokens ~\cite{pmlr-v235-jin24b,peysakhovich2023attentionsorting,xiao2024efficientstreaming}. Since these tokens  are crucial for continuous generation,  we introduce hyperparameters \(s_1\) and \(s_2\) to control the widths of fine-grained regions at the head and tail, respectively. The final relative position matrix is defined as:

\begin{equation}
\begin{aligned}
& PE[i][j] = \\
& \begin{cases}
    i-j, & i-j \leq s_1 \\ 
    \left\lfloor \dfrac{m-s_1-s_2}{l-s_1-s_2}(i-j-s_1) + s_1 \right\rfloor, & s_1 < i-j < l-s_2 \\
    m-l+i-j, & i-j \geq l-s_2 . \\ 
    \end{cases}
\end{aligned}
\end{equation}

The design rationale for each region reflects different modeling priorities: The head region  maintains fine-grained locality between neighboring tokens, which is crucial for continuous generation of next tokens. The middle region employs linear transformation with floor operation to preserve global relative ordering while supporting arbitrary input lengths without favoring specific positions. The tail region  preserves fine-grained relationships for distant tokens, considering that key instructions or questions often appear at sequence boundaries.

\paragraph{Attention Computation with Dynamic Mapping.}
We implement our dynamic mapping strategy within the RoPE-based LLM by defining region-specific position indices \(P_q[i]\) and \(P_k[j]\) that determine rotation matrices during attention computation:

\textbf{Head region} (\(i-j\le s_1\)): Standard relative positions for fine-grained local interactions:
\begin{equation}
    P_q^{(H)}[i]=i, \quad P_k^{(H)}[j]=j.
\end{equation}

\textbf{Middle region} (\(s_1 < i-j < l-s_2\)): Compressed group position indices via linear transformation:

\begin{equation}
    \begin{aligned}
    & P_q^{(M)}[i]=\left\lfloor \frac{m-s_1-s_2}{l-s_1-s_2}i + \frac{(l-m)s_1}{l-s_1-s_2} \right\rfloor, \\ 
    & P_k^{(M)}[j]=\left\lfloor  \frac{m-s_1-s_2}{l-s_1-s_2}j \right\rfloor.
    \end{aligned}
\end{equation}

\textbf{Tail region} (\(i-j \ge l-s_2\)): Shifted mapping for boundary context preservation:
\begin{equation}
    P_q^{(T)}[i]=m-l+i, \quad P_k^{(T)}[j]=j.
\end{equation}

Importantly, our dynamic mapping strategy preserves continuity across region boundaries, ensuring that the monotonicity property of relative position encoding is maintained: tokens at greater distances consistently receive larger relative position values. In addition, \method{} can seamlessly integrate with efficient attention implementations such as FlashAttention2~\cite{dao2023flashattention2}. The mathematical guarantee of the continuity property and the implementation pseudocode  of \method{} with FlashAttention2 are detailed in Appendix B and C, respectively.

\begin{table*}[!tb]
  \centering
   \small
 \resizebox{0.95\textwidth}{!}{
  \begin{tabular}{lcccccccccccccc}
    \toprule
\multirow{2}{*}{} & \multicolumn{7}{c}{\textbf{LongBench}}& \multicolumn{7}{c}{\textbf{L-Eval}} \\
\cmidrule(lr){2-8} \cmidrule(lr){9-15} 
&\textbf{S-DOC}&\textbf{M-DOC}&\textbf{Sum}&\textbf{ICL}&\textbf{Syn.}&\textbf{Code} &\textbf{Avg.}&\textbf{Coursera}&\textbf{GSM}&\textbf{QuALITY}&\textbf{TOEFL}&\textbf{Sfiction} &\textbf{Avg.}&\\
\midrule 
    Llama2-7B-Chat & 24.90 & 22.53 & 24.63 & 60.00 & 5.95 & 48.15 &31.52& 29.21 & 19.00 & 37.62 & 51.67 & 60.15 & 39.53 \\
    ~~+ SelfExtend(16K) & 27.30 & 26.23 & 24.81 & 64.15 & 5.74 & 57.46& 34.62& 35.76 & 25.00 & 41.09 & 55.39 & 57.81 & 43.01 \\
    ~~+ SelfExtend(25K) & 27.56 & 27.14 & 24.89 & 62.86 & 3.70 & 57.01& 34.30& 36.19 & 32.00 & 42.07 & 57.99 & 53.12 & 44.27 \\
    ~~+ DCA(16K) & 27.04   & 20.91 & 24.76  & 64.19  & 4.08 & 54.76 & 33.02& 32.12 & 32.00 & 35.14 & 57.62 & 61.72 & 43.72 \\
    ~~+ DCA(25K) & 26.29   & 19.90 & 24.85  & 63.18  & 4.36 & 54.16 & 32.48& 41.27 & 29.00 & 41.08 & 57.24 & 59.37 & 45.59   \\
     ~~+ YaRN(16K) &  22.77  & 13.51  & 21.83  & 47.91 & 1.98 & 47.68&26.08&  42.44  & 20.00  & 41.58  & 58.36 & 42.18 & 40.91 \\
    ~~+ YaRN(25K) & 22.72   & 25.92  & 25.56  & 60.26 & 2.24 & 46.85&31.35& 43.16   & 14.00  & 38.11  & 53.53 & 56.25 & 41.01 \\
    ~~+ NTK(16K) & 17.20 & 10.59 & 17.40 & 38.27 & 0.69 & 40.43 & 20.79 & 37.06 & 16.00& 33.66 & 61.71 & 40.62 & 37.81 \\
    ~~+ NTK(25K) & 19.58 & 13.90 & 21.63 & 47.56 & 3.89 & 42.40 & 25.03 & 39.53 & 11.00 & 29.70 & 50.92 & 48.43 & 35.91 \\
    \rowcolor{gray!30}
     ~~+ Ours & 29.70 & 27.45  & 25.82 & 63.93 & 4.41 & 55.72 & \textbf{35.07}& 42.58 & 35.00  & 42.57 & 57.24 & 63.28 & \textbf{48.13} \\
\midrule 
    Llama3-8B-Ins & 36.83 & 34.91 & 26.83 & 69.08 & 33.50 & 54.11 & 42.38& 52.76 & 79.00 & 59.40 & 80.95 & 66.40 & 67.07\\
     ~~+ SelfExtend(16K) & 14.76 & 7.26 & 21.53 & 48.40 & 2.11 & 57.14 & 24.71& 55.08 & 78.00 & 63.36 & 80.66 & 67.34 & 68.88 \\
     ~~+ SelfExtend(32K) & 38.57 & 33.98 & 28.55 & 68.59 & 46.06 & 37.15& 42.22& 56.39 & 79.00 & 64.35 & 80.66 & 66.56 & 69.39 \\
      ~~+ DCA(16K) & 39.10   & 38.77 & 27.44  & 63.54  & 37.75 & 64.95& 44.50& 56.10   & 75.00 & 61.38  & 81.41  & 73.43 & 69.46  \\
     ~~+ DCA(32K) & 40.16   & 38.86 & 27.83  & 63.53  & 37.50 & 64.59& 44.70&56.10 &77.00 & 59.40 & 81.41 & 75.78 & 69.93 \\
     ~~+ YaRN(16K) &  35.42  & 33.81  & 26.40  & 56.58 & 48.50 & 61.90 &42.34 & 58.43& 78.00 &65.34& 79.92 & 66.40 &69.61\\
     ~~+ YaRN(32K) & 38.74   & 42.71  & 28.91  & 59.23 & 47.75 & 65.04& 45.90 &56.54&76.00&67.32&79.92&74.21&70.79 \\
    ~~+ NTK(16K)&  32.59 & 21.81 &  22.17 & 46.50 & 26.12 & 57.67 & 33.55 & 51.45 & 76.00 & 63.36 & 80.66 & 53.12 & 64.91\\
     ~~+ NTK(32K)& 35.85 & 40.39 & 27.48 & 56.62 & 51.50 & 61.93 & 44.24 & 53.34 & 79.00 & 64.85 & 82.52 & 64.06 & 68.75 \\
     \rowcolor{gray!30}
     ~~+ Ours & 41.94 & 44.48  & 28.70 & 61.00 & 46.00 &65.73 & \textbf{46.99}& 60.02 & 80.00  & 66.33 & 79.93 & 72.65 &\textbf{71.78} \\
    \bottomrule
  \end{tabular}
  } 
   \caption{Performance comparison of different baselines on LongBench and L-Eval. The LongBench and L-Eval benchmarks consist of 16 and 5 tasks, respectively. The number (e.g. ‘16K’) indicates the maximum input length.}
    \label{tab:longbench}
\end{table*}

\begin{table*}[!htp]
  \centering
  \small
  \begin{tabular}{lccccccccc}
    \toprule
\textbf{Models}&\textbf{En.MC}&\textbf{En.QA}&\textbf{En.sum}&\textbf{Code.debug}&\textbf{Re.KV}&\textbf{Re.Number}&\textbf{Re.Passkey}&\textbf{Avg.} \\
\midrule 
    Llama3-8B-Ins & 50.66 & 10.54 &16.23 & 25.13  & 6.60 & 6.78 & 20.12  & 19.43  \\
     ~~+ SelfExtend~(32K) &  50.66 & 14.06 & 15.13 & 24.87  & 3.60 & 27.12 & 27.12 & 23.22  \\
     ~~+ DCA~(32K)  & 52.84 &  13.90 & 18.79 & 25.38  & 4.40 & 27.12 & 27.12 & 24.22\\
     ~~+ YaRN~(32K)  & 39.30 & 10.42 & 15.84 & 26.14  & 0 & 26.10 & 27.12 & 20.70 \\
     \rowcolor{gray!30}
     ~~+ Ours~(32K)  & 55.02 & 16.36 & 20.49 & 25.63  & 17.40 & 27.12 & 27.12 & \textbf{27.02}  \\
     \midrule 
     Llama3-8B-Ins  & 50.66 & 10.54 &16.23 & 25.13  & 6.60 & 6.78 & 20.12  & 19.43  \\
     ~~+ SelfExtend~(64K) & 53.71 & 15.10 & 15.22 & 21.57  & 2.80 & 54.24 & 54.24 & 30.98  \\
     ~~+ DCA~(64K) & 50.66 &14.35 & 18.98 & 24.11  & 2.00 &52.88 & 54.24 & 31.03  \\
     ~~+ YaRN~(64K)  & 5.24 & 1.63 & 12.06 & 0.51  & 0 & 3.56 & 24.96 & 6.85 \\
     \rowcolor{gray!30}
     ~~+ Ours~(64K) & 55.90 & 15.49 & 23.10 & 24.11 & 10.80 & 54.24 & 54.24 & \textbf{33.98} \\
\midrule 
    Llama3.1-8B-Ins & 67.25  & 14.57 & 25.42 & 22.08 & 54.80 & 99.49 & 100.00 & 54.80 \\
     ~~+ SelfExtend~(128K)   & 68.12 & 15.58 & 26.26 & 24.62 & 57.20 & 100.00 & 100.00 & 55.96\\
     ~~+ DCA~(128K) & 67.79 & 13.67 & 26.75 & 24.37 & 70.40 & 100.00 & 100.00 & 57.55\\
     ~~+ YaRN~(128K)  & 54.59 & 10.30 & 25.76 & 27.92 & 16.40 & 94.41 & 99.83 & 47.03 \\
     ~~+ STRING~(128K)  & 71.18 & 14.39 & 27.81 & 30.46 & 81.40 & 99.83 & 100.00 & 60.72  \\
     \rowcolor{gray!30}
     ~~+ Ours~(128K) & 70.30 & 19.51 & 28.54 & 29.19 & 92.60 &  99.83 & 100.00 & \textbf{62.85}\\
    \bottomrule
  \end{tabular}
   \caption{Results of Llama3-8B-Instruct and Llama3.1-8B-Instruct on $\infty$Bench. Since all evaluation inputs exceed 64K tokens, the results on Llama3 fully reflect the extrapolation performance of \method{} and other baseline methods.}
    \label{infinite}
\end{table*}

\section{Experiments}
\subsection{Experimental Setup}
\paragraph{Baselines.} We conduct experiments on Llama2-7B-Chat-4K~\cite{touvron2023llama2}, Llama3-8B-Instruct-8K~\cite{meta2024introducing} and Llama3.1-8B-Instruct-128K~\cite{dubey2024llama}. In addition to comparing \method{} with the original RoPE, we evaluate it against several commonly used length extrapolation baselines, including NTK-RoPE~\cite{NTK}, YaRN~\cite{peng2023yarnefficientcontextwindow}, SelfExtend~\cite{jin2024llmmaybelonglmselfextend}, DCA~\cite{an2024training}, and STRING~\cite{an2024does}. Among these, NTK-RoPE and YaRN  enable extrapolation by modifying the base. STRING  is designed to enhance performance within the original context window by modifying position indices. We reproduced their results using scripts from their official repositories, with configurations aligned to those reported in their original papers. All experiments are performed on a single A800 GPU using BF16 precision and  accelerated with FlashAttention2~\cite{dao2023flashattention2}.
\paragraph{Datasets.} We conduct comprehensive evaluations of our proposed method on five widely recognized long-context benchmarks. 

\textbf{LongBench \& L-Eval:} These two benchmarks are widely used for evaluating long-context methods, covering both real-world tasks and synthetic tasks. LongBench consists of 16 subtasks. For L-Eval, we select five representative tasks from it for evaluation. The average input length for both datasets is below 32K tokens.

\textbf{$\infty$Bench:} Designed to assess ultra-long-context understanding, $\infty$Bench includes a diverse set of real-world tasks and synthetic tasks with an average input length of more than 128K tokens.

\textbf{RULER:} A synthetic benchmark for long-context evaluation with 13 complex tasks across 4 categories: 8 NIAH variants, long-context QA, variable tracking, and counting. A key advantage of RULER is its flexibility: task length can be customized based on the model's context window, allowing evaluations up to 128K tokens or beyond.

\textbf{PG-19:} PG-19 is a traditional benchmark for long-context language modeling comprising over 28,000 books published before 1919, serves as the final dataset in our experiments.  Evaluations on PG-19 can be found in Appendix D.

\begin{figure*}[!tb]
    \centering
    \includegraphics[width=1\linewidth]{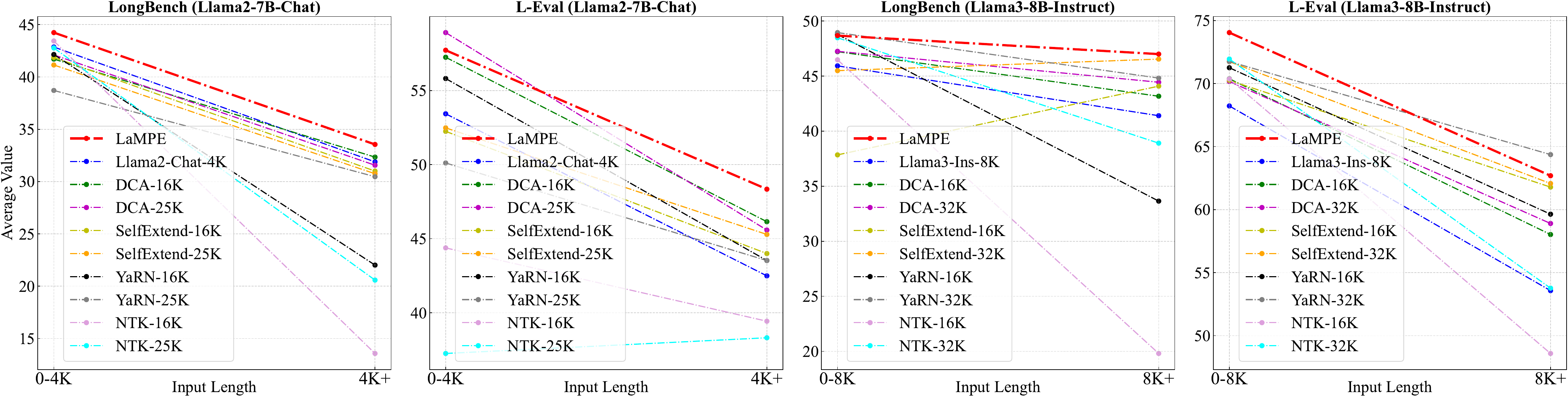}
    \caption{Performance statistics across different input length intervals for LongBench and L-Eval datasets. For Llama2-7B-Chat, input lengths are divided into [0–4K] and [4K+], while for Llama3-8B-Instruct, the intervals are [0–8K] and [8K+].}
    \label{fig:diff_len}
\end{figure*}

\subsection{Performance on Long Context Benchmarks}
\paragraph{LongBench \& L-Eval.} As shown in Table \ref{tab:longbench}, \method{} achieves the best overall performance on both LongBench and L-Eval with Llama2-7B-Chat and Llama3-8B-Instruct. Since baseline methods require manual tuning of extrapolation factors or group sizes, there exists an upper bound for extrapolation. However, as the distribution of datasets does not consist entirely of the specified length, we evaluate two different maximum input lengths under two different experimental settings. Specifically, we set maximum input lengths of 16K and 25K tokens for Llama2, and 16K and 32K tokens for Llama3. In contrast, our method does not require manual setting of maximum length and only needs one set of experiments.

In summary, our method applied to Llama2 and Llama3 respectively outperforms the best baseline by 0.45 and 1.09 points on average across 16 tasks on LongBench. On L-Eval, our method achieves improvements of 2.54 and 0.99 points, respectively.  Specifically, on Llama2-7B-Chat, position indices modified methods (SelfExtend, DCA) significantly outperform base-modified methods (YaRN, NTK-RoPE), though this performance gap narrows on Llama3-8B-Instruct due to the model's stronger capacity.

\paragraph{$\infty$Bench.} We evaluate the performance of various baselines on Llama3-8B-Instruct at extrapolation lengths of 4× (32K) and 8× (64K). Since all tasks in the $\infty$Bench  greatly exceed 64K tokens, these results directly reflect the extrapolation capability of each method. The experimental results are shown in Table~\ref{infinite}, \method{} consistently achieves higher average scores than the baselines at both 32K and 64K. We observe that methods based on modifying position indices tend to be more stable than those modifying base. Notably, when the extrapolation length reaches 8×, YaRN exhibits a substantial performance degradation.  In addition to evaluating \method{} on Llama3-8B-Instruct with an 8K context window, we further apply it to Llama3.1-8B-Instruct, which supports a 128K context window. As shown in the third block of Table~\ref{infinite}, when the maximum input length is set to 128K tokens, our method again achieves the best performance. In particular, for the KV retrieval task, our method outperforms the original RoPE by 37.8 points. It indicates that \method{} can also effectively scale to long-context LLMs.

\begin{figure}[!htb]
    \centering
    \includegraphics[width=0.98\linewidth]{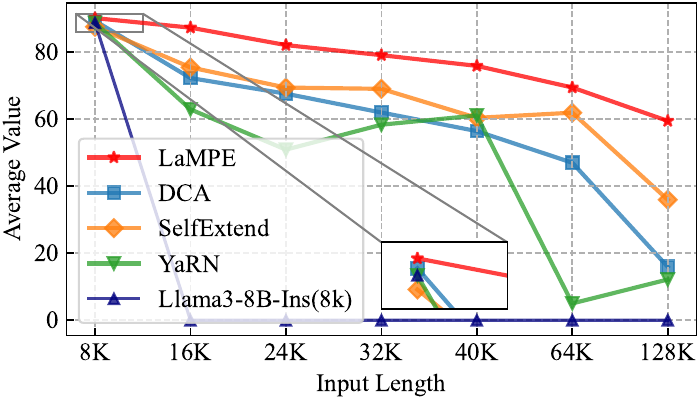}
    \caption{Performance of Llama3-8B-Instruct on the RULER benchmark across input lengths from 8K to 128K.}
    \label{fig:ruler}
\end{figure}

\paragraph{RULER.} Figure~\ref{fig:ruler} presents the performance of  \method{}  under stress testing on the RULER benchmark. The evaluation spans from the pretraining context window of 8K up to 128K. Across this entire range, \method{} consistently achieves the highest accuracy compared to all other baseline methods. Notably, we found similar patterns as observed on $\infty$Bench: when the extrapolation length reaches 64K, YaRN exhibits a dramatic performance drop. In contrast, modifying position indices methods (SelfExtend, DCA) are relatively stable, indicating that YaRN's extrapolation upper bound is lower than that of modifying position indices methods. Furthermore, besides achieving the best performance at extrapolation lengths, our method also achieves 90.57 within the pretraining context window, surpassing the original RoPE's 88.76. This demonstrates that utilizing the left-skewed frequency distribution of relative positions can achieve efficient extrapolation. Detailed per-task scores are available in Appendix E.

\subsection{Analysis}
\paragraph{Performance Enhancement on Pretraining Context Window.} To further assess \method's performance within the pretraining context window, we categorize LongBench and L-Eval based on whether they fall inside or exceed the pretraining context window, and compute performance metrics for the two categories separately. As shown in Figure~\ref{fig:diff_len}, our method consistently achieves optimal or suboptimal performance both at extrapolation lengths and within the pretraining context window. This demonstrates that \method{} can serve as a general technique for long-context scaling. Moreover, it can also be observed from the figure that DCA demonstrates strong stability across both extrapolation lengths and the pretraining context window, while SelfExtend's results on Llama3 in LongBench show that it impairs performance within the original window, which also highlights the limitations of manually setting group size.

\begin{figure}[!htb]
    \centering
    \includegraphics[width=0.98\linewidth]{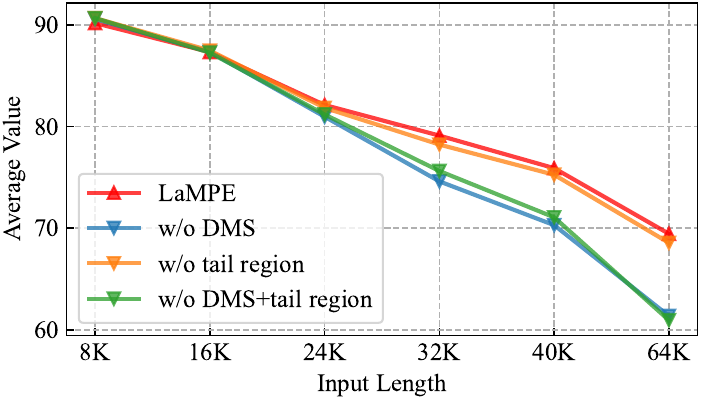}
    \caption{Performance comparison of different ablation methods on RULER benchmark using Llama3-8B-Instruct.}
    \label{fig:ruler_abla}
\end{figure}

\paragraph{Ablation Study.} 
To evaluate the contributions of two components in \method{}, we introduce two variants for the ablation study: (1) Ours w/o DMS, which uses a fixed mapping length that remains constant across inputs of varying lengths; (2) Ours w/o tail region, which doesn't recover long-range dependencies between current tokens and initial tokens. As shown in Figure~\ref{fig:ruler_abla}, we observe that the dynamic mapping strategy significantly improves LLM performance. Specifically, without the dynamic mapping strategy, the performance drops  as input length increases, resulting in a noticeable gap compared to \method{}. Additionally, removing the tail region also leads to a performance drop, though the impact is smaller than that of removing the dynamic mapping strategy. These results demonstrate the necessity of both components, with the dynamic mapping strategy being particularly critical for maintaining performance across varying input lengths.

\begin{figure}[!htb]
    \centering
    \includegraphics[width=0.98\linewidth]{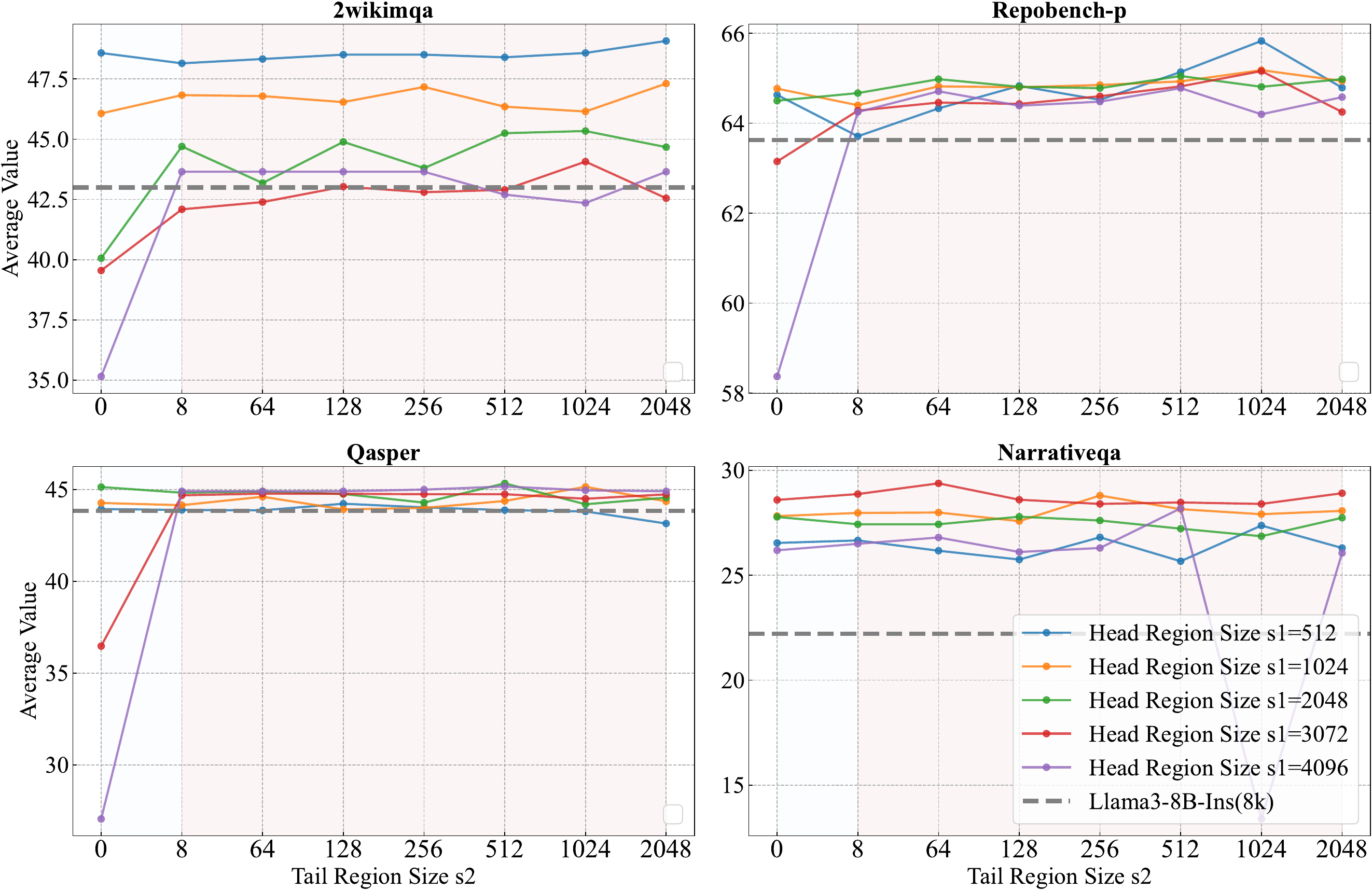}
    \caption{Ablation study results on LongBench using different head region and tail region sizes.}
    \label{fig:head_tail_abla}
\end{figure}

Furthermore, we conduct experiments on four real-world datasets from LongBench to examine the impact of two main hyperparameters in \method{}: the head region size $s_1$ and the tail region size $s_2$.  As illustrated in Figure~\ref{fig:head_tail_abla}, our results demonstrate that performance remains robust when both head and tail region sizes are  smaller than the mapping length. To evaluate extreme cases with large head region sizes, we increase the mapping length correspondingly to match the head region size. Our findings reveal significant performance degradation when long-range dependencies are not recovered ($s_2 = 0$), particularly on tasks such as 2wikimqa, Hotpotqa, and Repobench-p. However, incorporating even a minimal tail region ($s_2 = 8$) effectively restores performance, demonstrating the effectiveness of our method in recovering dependencies between current tokens and initial tokens. In summary, optimal performance is achieved with a head region size of approximately 1/8 to 1/16 of the pretraining context window, paired with a small but non-zero tail region (e.g., 8 to 1024).

\section{Related Work}
\paragraph{Positional Encoding.} Positional encoding is a fundamental technique for incorporating position information into transformers and can be broadly categorized into two types: absolute and relative positional encoding. Absolute positional encoding associates each position with either learnable parameters~\cite{bert,Lan2020ALBERT} or fixed sinusoidal embeddings as employed in the original transformer~\cite{vaswani-nips-2017-attention}. In contrast, relative positional encoding encodes the relative distances between tokens rather than their absolute positions. Among relative positional encoding methods, ALiBi~\cite{press2022train} introduces a position-biased attention mechanism that applies linear decay to attention weights based on token distance, successfully implemented in MPT~\cite{MosaicML2023Introducing}. Another notable method is Rotary Position Embedding (RoPE)~\cite{su2024roformer}, which has achieved widespread adoption due to its excellent performance. RoPE has been integrated into several state-of-the-art language models~\cite{codellama,dubey2024llama,jiang2023mistral7b}, demonstrating its effectiveness across diverse architectures and applications. Our work focuses on RoPE-based LLMs, aiming to enhance their capability for long-context scaling.

\paragraph{Length Extrapolation based RoPE.} We divide length extrapolation methods based RoPE on into two  types: base-modified and position indices modified. Inspired by neural tangent kernel theory, NTK-aware~\cite{NTK}, Dynamic-NTK~\cite{dyn-ntk} and YaRN~\cite{peng2023yarnefficientcontextwindow} perform extrapolation on high-frequency components and interpolation on low-frequency components. Subsequently, LongRoPE\cite{longrope,longrope2} uses an evolutionary search to exploit two forms of  non-uniformities in RoPE Frequency. These methods can work without training, but a small number of training steps can lead to better performance. In contrast, ReRoPE\cite{kexuefm-9708}, SelfExtend\cite{jin2024llmmaybelonglmselfextend} and DCA\cite{an2024training} achieve length extrapolation by modifying position indices that exceed a certain range to the limiting value.  The first type methods are orthogonal to our method and could be integrated with our techniques. Our method belongs to the second type and employs a dynamic mapping strategy to fully leverage effective positions for length extrapolation.

\section{Conclusion}
We have proposed \method{}, a training-free method for adaptive long-context scaling of RoPE-based large language models.  \method{} models the dynamic relationship between mapping length and input length using a parametric scaled sigmoid function, and incorporates a multi-grained attention mechanism to jointly capture fine-grained locality and long-range dependencies.  Extensive experiments across long-context  benchmarks demonstrate that LAMPE significantly improves extrapolation performance and enhances performance within the original context window.

\bibliography{aaai2026}
\clearpage
\onecolumn
\section*{Appendix}

\subsection*{A. Visualization of two patterns of PPL variation with mapping length}
\begin{figure*}[!htb]
    \centering
    \includegraphics[width=0.78\linewidth]{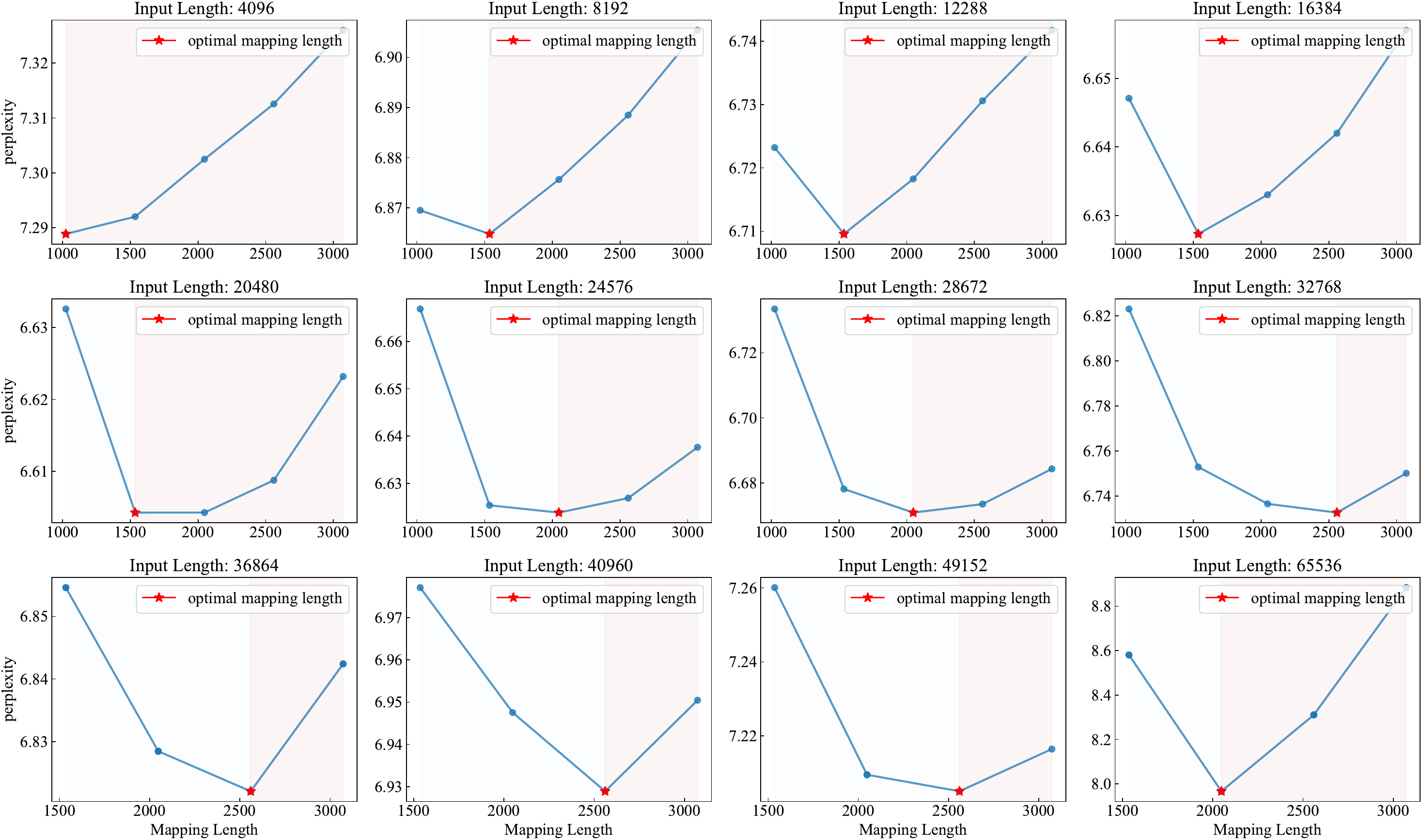}
    \caption{Visualization of two  patterns of PPL variation with mapping length in Llama2-7B-Chat.}
    \label{fig:ppl_change2}
\end{figure*}

\begin{figure*}[!htb]
    \centering
    \includegraphics[width=0.78\linewidth]{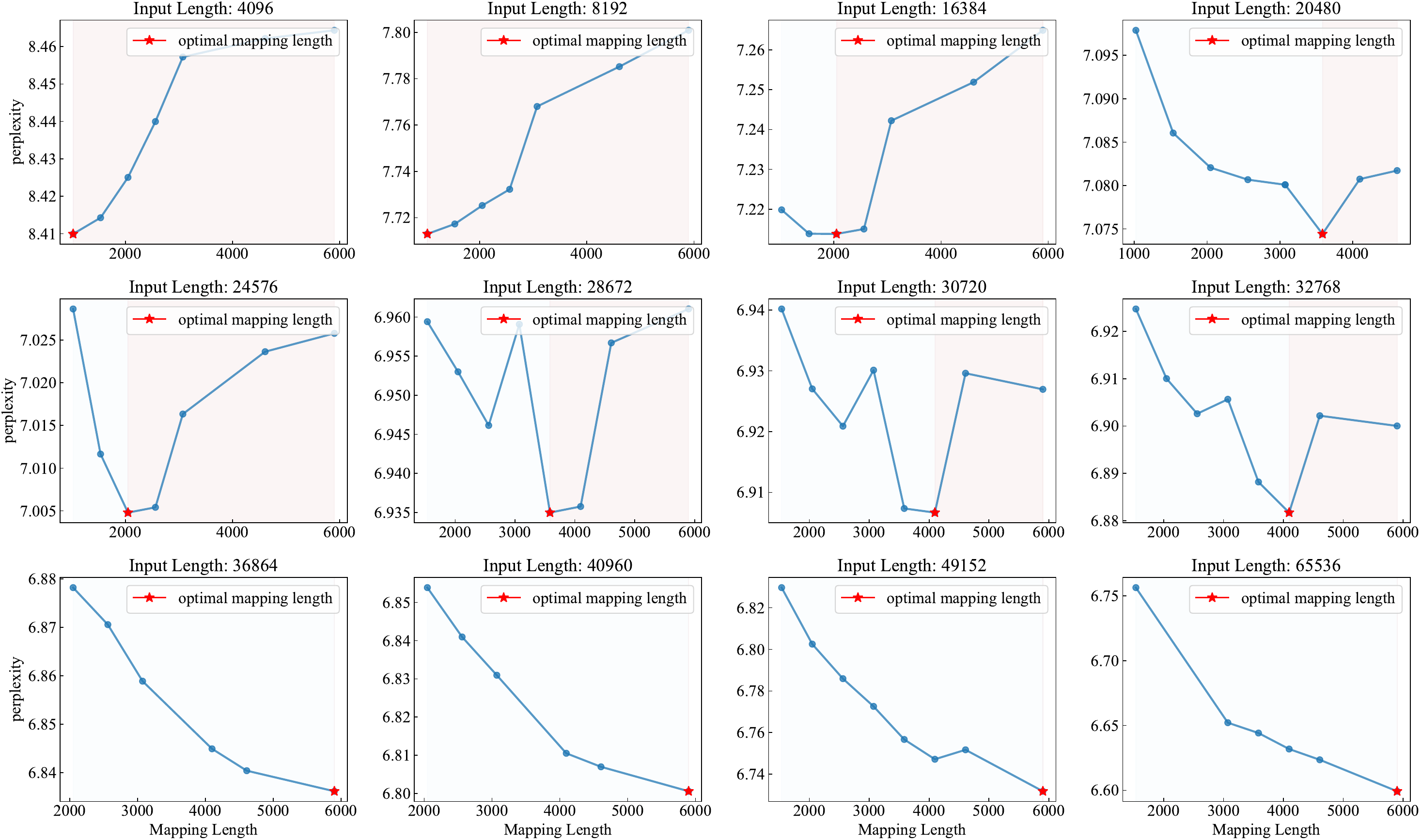}
    \caption{Visualization of two  patterns of PPL variation with mapping length in Llama3-8B-Instruct.}
    \label{fig:ppl_change3}
\end{figure*}

The V-shaped and monotonically decreasing patterns for Llama2 and Llama3 are illustrated in Figures~\ref{fig:ppl_change2} and  \ref{fig:ppl_change3}. As the input length increases, the optimal mapping length increases correspondingly. We set the maximum mapping length to 3/4 of the pretraining context window, as beyond this point perplexity  increases substantially. This observation aligns closely with findings from STRING~\cite{an2024does}, where needle-in-a-haystack tasks consistently fail in the last 1/3 of sequences, occurring predominantly in the tail of the position frequency distribution. These results demonstrate that mechanically using the entire pretraining context window for extrapolation is inadequate and that one should instead fully leverage well-trained positions.

\subsection*{B. Mathematical Guarantee of the Continuity Property}

\method{} achieves length extrapolation by mapping larger position indices to smaller ones through region-specific transformations. This section provides a rigorous mathematical proof that the dynamic mapping strategy preserves the monotonicity property essential for maintaining coherent relative position relationships in attention mechanisms.

We first formalize the monotonicity property for relative position encodings. Let \(PE[i][j]\) denote the relative position value  between token \(i\) and token \(j\), where \(PE[i][j] = P_q[i]-P_k[j]\) represents the position difference that determines the rotation angles in RoPE-based attention computation.

\textbf{Definition (Monotonicity Property):} For any token position \(i\)
, the relative position encodings satisfy monotonicity if for all \(0 \le j_1 < j_2 \le i\), we have \(PE[i][j_1] \ge PE[i][j_2]\).

This property ensures that tokens closer to the current position maintain smaller relative distances, preventing discontinuous jumps in the relative position sequence that could disrupt attention patterns.

To establish our proof, we introduce the following notation from the middle region transformation:

\begin{equation}
k = \frac{m-s_1-s_2}{l-s_1-s_2}, \quad b = \frac{(l-m)s_1}{l-s_1-s_2},
\end{equation}
where \(0 < k \le 1\) since \(0 < m\le l\) in our method. These parameters satisfy two properties:
\begin{equation}
    ks_1 + b = s_1, \quad k(l-s_2) + b = m - s_2 .
\end{equation}

Our proof  relies on the floor function inequality:
\begin{equation}
    \lfloor x \rfloor - \lfloor y \rfloor - 1 \le \lfloor x - y\rfloor \le \lfloor x \rfloor - \lfloor y \rfloor .
\end{equation}

Since \method{} partitions the sequence into three regions with distinct mapping functions, we establish monotonicity by proving continuity at the boundaries between adjacent regions, as monotonicity within each region follows directly from the mapping definitions.

\textbf{Boundary between the Head and Middle Regions:} Consider any token position at
\(i\) and boundary positions \(j_1\) and \(j_2\) where \(i-j_1=s_1\) (head region) and \(i-j_2 = s_1 + 1\) (middle region). We need to show \(PE[i][j_2] \ge PE[i][j_1]\). Note that \(PE[i][j_1] = i-j_1 = s_1\) by the head region mapping, while \(PE[i][j_2]\) follows the middle region transformation.
\begin{align}
    PE[i][j_2] & = \lfloor ki+b \rfloor - \lfloor kj_2\rfloor \ge \lfloor k(i-j_2) + b \rfloor,  \\
    & = \lfloor (ks_1+b) +k \rfloor = s_1 +\lfloor k \rfloor, \\
    & \ge s_1 = PE[i][j_1].
\end{align}

\textbf{Boundary between the Middle and Tail Regions:} Consider boundary positions where \(i-j_2=l-s_2-1\) (middle region) and \(i-j_3=l-s_2\) (tail region). We need to show \(PE[i][j_2] \le PE[i][j_3]\). Note that \(PE[i][j_3] = (m-l+i)-j_3 = m-s_2\) by the tail region mapping.
\begin{align}
    PE[i][j_2]  & = \lfloor ki+b \rfloor - \lfloor kj_2\rfloor \le \lfloor k(i-j_2) + b \rfloor + 1, \\
    & = \lfloor k(l-s_2)+b+1-k \rfloor = m-s_2+\lfloor 1-k\rfloor, \\
    & = m-s_2 =PE[i][j_3] .
\end{align}

These boundary conditions, combined with the inherent monotonicity within each region, establish that our dynamic mapping strategy preserves the monotonicity property across the entire sequence. This mathematical guarantee ensures that \method{} maintains coherent relative position relationships during length extrapolation, preventing the attention mechanism from encountering discontinuous position patterns that could degrade model performance.

\subsection*{C. Pseudocode of \method{} with FlashAttention2}
As shown in Algorithm~\ref{alg:method}, we provide a Python-style pseudocode implementation of our method integrated with FlashAttention2. The implementation consists of three distinct regions, each with specific attention mechanisms:

\textbf{Head Region.} As shown in Algorithm~\ref{alg:method} \texttt{(line 5-8)}, there is no need to modify the query or key. This is a standard sliding window attention with a window size of $s_1$.

\textbf{Middle Region.} \texttt{Line 10-14} corresponds to the implementation of the linear normalization. It is necessary to modify the position indices of query and key to the normalized query \texttt{(line 12)} and key \texttt{(line 13)}. We matmul the last query vector \texttt{Q\_m[s1-$l$:]} and key vector  \texttt{K\_m[:$l$-s1]} with sliding window size $l-s_1-s_2$.

\textbf{Tail Region.} The tail region implements full attention using a lower triangular mask, where the query vectors are modified with position encoding \texttt{(line 18)} while the key vectors remain unchanged \texttt{(line 19)}. This step computes the  attention for the last $s_2$ tokens \texttt{(line 20)}.

\definecolor{customTeal}{RGB}{0, 128, 128} 
\definecolor{emphasisColor}{RGB}{255, 0, 0} 
\lstset{
    language=Python,         
    basicstyle=\fontsize{7.0pt}{8.0pt}\ttfamily\selectfont,
    keywordstyle=\color{customTeal},    
    stringstyle=\color{customTeal},    
    commentstyle=\color{customTeal},     
    morecomment=[l][\color{magenta}]{\#},
    breaklines=true,                
    showstringspaces=false,
    escapeinside={(*@}{@*)}, %
    numbers=left,          
    stepnumber=1,           
    numberstyle=\tiny\color{gray}, 
    numbersep=5pt,         
    xleftmargin=1.5em,      
    frame=none,              
}
\begin{algorithm}[H]
\caption{\footnotesize Pseudocode of \method{} with FlashAttention2} \label{alg:method}
\begin{lstlisting}[language=Python]
#  Q, K, V: Queries, Keys, Values
#  s1, s2: head region size, tail region size
#  m, l: mapping length ,input  length
#  Head Region
pos = [0,1,2,...(*@$l$@*)-1] 
Q_h = apply_rotary_pos_emb(Q, pos)
K_h = apply_rotary_pos_emb(K, pos)
O_h, lse_h = flash_attn_func(Q, K, window_size=(s1, 0))
#  Middle Region
q_m_pos = floor((*@$\frac{m-s1-s2}{l-s1-s2}$@*) * pos + (*@$\frac{(l-m)s1}{l-s1-s2}$@*)) 
k_m_pos = floor((*@$\frac{m-s1-s2}{l-s1-s2}$@*) * pos)
Q_m = apply_rotary_pos_emb(Q, q_m_pos)
K_m = apply_rotary_pos_emb(K, k_m_pos)
O_m, lse_m = flash_attn_func(Q_m[s1-(*@$l$@*):], K_m[:(*@$l$@*)-s1], window_size=((*@$l$@*)-s1-s2, 0))
#  Tail Region
q_t_pos = pos +m - (*@$l$@*)
k_t_pos = pos
Q_t = apply_rotary_pos_emb(Q, q_t_pos)
K_t = apply_rotary_pos_emb(K, k_t_pos)
O_t, lse_t = flash_attn_func(Q_t[-s2:], K_t[:s2], window_size=(-1, -1)) 
# Merge hidden_state O_h, O_m, O_t
# Step 1: Split outputs
head_h = O_h[:, :s1]
midd_h = O_h[:, s1:(*@$l$@*)-s2-s1]
tail_h = O_h[:, (*@$l$@*)-s2:]
midd_m = O_m[:, :(*@$l$@*)-s2-s1]
tail_m = O_m[:, (*@$l$@*)-s2-s1:]
# Step 2: Merge middle parts
gate1 = sigmoid(lse_m[:(*@$l$@*)-s2-s1] - lse_h[s1:(*@$l$@*)-s2])
gate2 = 1 - gate1
midd = midd_h * gate2 + midd_m * gate1
# Step 3: Merge tail parts
gate1 = 1 / (1 + exp(lse_h[(*@$l$@*)-s2:] - lse_m[(*@$l$@*)-s2-s1:]) + exp(lse_t - lse_m[(*@$l$@*)-s2-s1:]))
gate2 = 1 / (1 + exp(lse_m[(*@$l$@*)-s2-s1:] - lse_t) + exp(lse_h[(*@$l$@*)-s2:] - lse_t))
gate3 = 1 / (1 + exp(lse_t - lse_h[(*@$l$@*)-s2:]) + exp(lse_m[(*@$l$@*)-s2-s1:] - lse_h[(*@$l$@*)-s2:]))
tail = tail_m * gate1 + O_t * gate2 + tail_h * gate3
# Step 4: Concatenate
output = concat(head_h, midd, tail)
\end{lstlisting}
\end{algorithm}

\subsection*{D. Additional Experiment on PG-19}

\begin{table}[!htb]
  \centering
  \small
  \begin{tabular}{lcccccccc}
    \toprule
\multirow{2}{*}{\textbf{Method}} & \multicolumn{7}{c}{Evaluation Context Window Size}\\
\cmidrule(lr){2-9} 
& \textbf{4K} & \textbf{8K} & \textbf{16K} & \textbf{24K} & \textbf{32K} & \textbf{40K}& \textbf{64K} & \textbf{Avg.} \\ 
\midrule 
    Llama2-7B-Chat & 7.13 &$>10^2$ & $>10^2$& $>10^2$ & $>10^2$ & $>10^2$ & $>10^2$ & - \\
    ~~+ ReRoPE & 7.13 & 6.86 & 8.85 & 34.05 & 95.74 & $>10^2$ & $>10^2$ & - \\
     ~~+ SelfExtend & 7.13 & 6.85 &7.44& 22.72 & 74.38 & $>10^2$ & $>10^2$  & -\\
     ~~+ DCA & 7.12 & 6.87 & 6.79 &  6.87 & 7.07 & 7.36 & 8.53 & 7.23\\
     ~~+ YaRN & 7.13 & 6.96 & 7.68 & 7.74 & 8.40 & 10.68 & 30.90  & 11.36 \\
     ~~+ NTK   & 7.13 & 7.05 & 14.79 & $>10^2$ & $>10^2$ & $>10^2$&  $>10^2$ & - \\
     \rowcolor{gray!30}
     ~~+ Ours  & 7.29 & 6.86 & 6.62 & 6.62 & 6.73 & 6.97 & 7.96 & \textbf{7.00}\\
\midrule 
    Llama3-8B-Instruct & 8.81 & 8.14 & 39.67 & $>10^2$& $>10^2$ & $>10^2$& $>10^2$ & -  \\
    ~~+ ReRoPE & 8.81 & 8.18 & 7.78 & 7.65 & 7.62 & 7.62 & 7.63  & 7.89\\
     ~~+ SelfExtend & 8.81 & 8.10 & 7.56 & 7.34 & 7.26 & 7.16 & 7.00 & 7.60\\
     ~~+ DCA & 8.81 & 8.11 & 7.42 &7.01& 6.94 & 6.86  & 6.84   & 7.43  \\
     ~~+ YaRN & 8.81 & 8.14 & 7.28 &7.02&6.87 & 6.81 & 6.93 & 7.41\\
     ~~+ NTK & 8.81 &8.14 &7.99 & 7.83 & 7.99 & 8.29 & 9.76 & 8.40 \\
     \rowcolor{gray!30}
     ~~+ Ours & 8.41 & 7.70& 7.21 & 7.00 & 6.90&  6.80 &6.59& \textbf{7.23}\\
    \bottomrule
  \end{tabular}
   \caption{Perplexity (PPL) evaluation comparison of different baselines on PG-19.}
     \label{tab:ppl}
\end{table}

\begin{table}[!tb]
  \centering
  \begin{tabular}{lcccc}
    \toprule
\multirow{2}{*}{} & \multicolumn{4}{c}{\textbf{Pretraining Context Window (8K)}}  \\
\cmidrule(lr){2-5}
&\textbf{RoPE}&\textbf{SelfExtend}&\textbf{DCA}&\textbf{\method{}}\\
\midrule 
NIAH\_S1 & 100.00 & 100.00 & 100.00 & \cellcolor{gray!30}100.00 \\
NIAH\_S2 & 99.80  & 99.80  & 100.00 & \cellcolor{gray!30}100.00 \\
NIAH\_S3 & 99.80  & 100.00 &  99.40 & \cellcolor{gray!30}100.00 \\
NIAH\_M1 & 95.00  &  93.60 & 96.80  & \cellcolor{gray!30}98.60   \\
NIAH\_M2 & 84.60  & 89.80  & 95.60  & \cellcolor{gray!30}98.80   \\
NIAH\_M3 & 95.00  &92.60 & 94.60    & \cellcolor{gray!30}93.80   \\
NIAH\_MQ & 99.45  & 99.40 & 99.35   & \cellcolor{gray!30}99.35   \\
NIAH\_MV &  99.20 & 94.60  & 98.75  & \cellcolor{gray!30}99.75   \\
VT       &  91.08 & 82.40  &  88.84 & \cellcolor{gray!30}90.07   \\
CWE      &  92.59 & 92.13  & 91.19  & \cellcolor{gray!30}86.19   \\
FWE      &  83.40 & 81.60  & 83.33  & \cellcolor{gray!30}83.06    \\
QA\_1    &  66.40 & 65.80 & 67.20   & \cellcolor{gray!30}75.20   \\
QA\_2    &  47.60 & 47.00 & 46.60   & \cellcolor{gray!30}52.60   \\
\bottomrule
AVg.     &  88.76 & 87.59 & 89.35   & \cellcolor{gray!30}\textbf{90.57}  \\
\bottomrule
  \end{tabular}
   \caption{Comparison of per-task performance across different baselines on RULER within the pretraining context window.}
   \label{tab:ruler_8K}
\end{table}

\paragraph{PG19}  Due to limited computational resources,  we sample 200 paragraphs from 100 books to construct our test set, while keeping all other experimental settings consistent with prior work. This setup significantly reduces resource requirements while  providing a reliable assessment of LLMs' perplexity.  Table~\ref{tab:ppl} presents the perplexity  scores on the PG19,  \method{} maintains relatively low PPL growth even as input length increases. Compared to other  methods, \method{} consistently achieves either the best or second-best results. For Llama2-7B-Chat, \method{}  keeps PPL below 7 across the 8K–40K input length, while other methods like SelfExtend, YaRN, and NTK-RoPE all exceed 10. For Llama3-8B-Instruct, \method{} consistently achieves the lowest PPL across all input lengths. Furthermore, even within the original context window (8K), \method{} delivers meaningful gains, reducing PPL by approximately 0.4 points.

\begin{table*}[!tb]
  \centering
  \small
 \resizebox{0.95\textwidth}{!}{
  \begin{tabular}{lcccccccccccccc}
    \toprule
\multirow{2}{*}{} & \multicolumn{4}{c}{\textbf{16K}} & \multicolumn{4}{c}{\textbf{24K}} & \multicolumn{4}{c}{\textbf{32K}} \\
\cmidrule(lr){2-5} \cmidrule(lr){6-9} \cmidrule(lr){10-13} 
&\textbf{SelfExtend}&\textbf{DCA}&\textbf{YaRN}&\textbf{\method{}}&\textbf{SelfExtend}&\textbf{DCA}& \textbf{YaRN}&\textbf{\method{}}&\textbf{SelfExtend}&\textbf{DCA} &\textbf{YaRN} &\textbf{\method{}}\\
\midrule 
NIAH\_S1 & 100.00 & 100.00 & 100.00 & \cellcolor{gray!30}100.00 & 100.00 & 100.00 & 92.60 & \cellcolor{gray!30} 100.00& 100.00 & 100.00 & 91.20 & \cellcolor{gray!30}100.00 \\
NIAH\_S2 & 99.80  & 83.40  & 99.40 & \cellcolor{gray!30}100.00 & 98.40 & 74.60 & 94.80 & \cellcolor{gray!30} 100.00 & 99.80  & 56.50  & 95.40 & \cellcolor{gray!30}100.00 \\
NIAH\_S3 & 99.80  & 92.60 &  98.40 & \cellcolor{gray!30}99.80  & 99.80 & 84.60 & 92.60 & \cellcolor{gray!30} 100.00 & 100.00 & 71.80  & 90.20 & \cellcolor{gray!30}99.80  \\
NIAH\_M1 & 86.00  &  56.80 & 83.00  & \cellcolor{gray!30}96.60  & 83.00 & 52.60 & 48.20 & \cellcolor{gray!30} 95.00& 89.60  & 46.40  & 48.60 & \cellcolor{gray!30}92.80  \\
NIAH\_M2 & 48.80  & 47.80  &  24.00  & \cellcolor{gray!30}94.80 & 36.80 & 42.60 & 0.6 & \cellcolor{gray!30} 83.80 & 42.00  & 37.00  & 12.80 & \cellcolor{gray!30}81.40  \\
NIAH\_M3 & 37.80  &33.40   & 10.40   & \cellcolor{gray!30}73.40  & 21.20 & 29.20 & 22.40 & \cellcolor{gray!30}59.80 & 25.80  & 15.80  & 0.20  & \cellcolor{gray!30}56.80  \\
NIAH\_MQ & 99.25  & 92.10  & 97.85   & \cellcolor{gray!30}98.85 & 97.75 & 83.45 & 72.00 & \cellcolor{gray!30}98.10  & 98.40  & 81.80  & 94.80 & \cellcolor{gray!30}98.05 \\
NIAH\_MV &  93.10 & 89.20  & 97.95  & \cellcolor{gray!30}98.05  & 94.60 & 84.25 & 89.85 & \cellcolor{gray!30}98.40 & 94.95  & 77.30  & 91.35 & \cellcolor{gray!30}97.20  \\
VT       &  68.64 & 64.40  &  2.52 & \cellcolor{gray!30}83.00  & 52.99 &  64.32& 0 & \cellcolor{gray!30} 84.20& 60.12  & 65.44  & 76.28 & \cellcolor{gray!30}77.12  \\
CWE      &  54.28 & 87.41  & 14.02  & \cellcolor{gray!30}77.63  & 46.11 & 80.95 & 11.00 & \cellcolor{gray!30}56.72 & 19.40  & 68.31  & 0.04  & \cellcolor{gray!30}34.33  \\
FWE      &  91.13 & 92.20  & 87.26  & \cellcolor{gray!30}91.93 & 81.33 & 85.00 & 67.93 & \cellcolor{gray!30} 82.86 & 89.00  & 92.66  & 84.60 & \cellcolor{gray!30}90.00   \\
QA\_1    &  59.40 & 59.00 & 71.00  & \cellcolor{gray!30}73.20  & 51.40 & 54.60 & 38.80 & \cellcolor{gray!30} 64.20& 47.80  & 55.80  & 42.20 & \cellcolor{gray!30}57.00 \\
QA\_2    &  42.80 & 41.40 & 32.40   & \cellcolor{gray!30}48.00 & 34.90 & 41.80 & 31.20 & \cellcolor{gray!30} 44.60 & 38.00  & 36.80  & 31.00 & \cellcolor{gray!30}44.00 \\
\midrule 
AVg.     &  75.44 & 72.28 & 62.93   & \cellcolor{gray!30} \textbf{87.32} & 69.44 & 67.53 & 50.92 & \cellcolor{gray!30}\textbf{82.12}  & 69.06  & 61.99  & 69.06 & \cellcolor{gray!30}\textbf{79.12} \\
\midrule 
\multirow{2}{*}{} & \multicolumn{4}{c}{\textbf{40K}} & \multicolumn{4}{c}{\textbf{64K}} & \multicolumn{4}{c}{\textbf{128K}} \\
\cmidrule(lr){2-5} \cmidrule(lr){6-9} \cmidrule(lr){10-13} 
&\textbf{SelfExtend}&\textbf{DCA}&\textbf{YaRN}&\textbf{\method{}}&\textbf{SelfExtend}&\textbf{DCA}& \textbf{YaRN}&\textbf{\method{}}&\textbf{SelfExtend}&\textbf{DCA} &\textbf{YaRN} &\textbf{\method{}}\\
\midrule 
NIAH\_S1 & 100.00  &  99.60 & 99.00  & \cellcolor{gray!30}100.00   & 100.00 & 99.00 & 0 & \cellcolor{gray!30}100.00 & 100.00  & 85.20  & 75.40  & \cellcolor{gray!30}100.00\\
NIAH\_S2 & 98.20  &  50.00 & 93.20  & \cellcolor{gray!30}99.60 & 99.20  & 43.20 & 0 & \cellcolor{gray!30}98.80 & 50.40  & 31.20  & 11.80  & \cellcolor{gray!30}85.00\\
NIAH\_S3 & 91.60  & 61.40  &  98.40 & \cellcolor{gray!30} 99.80   & 98.40  & 44.00 & 0 & \cellcolor{gray!30}100.00&  8.20 &  20.20 &  0 & \cellcolor{gray!30}99.00\\
NIAH\_M1 & 72.40  &  45.00 &  60.00 & \cellcolor{gray!30}  91.00    & 75.80  & 37.80 & 0 & \cellcolor{gray!30}89.80& 29.60  &  21.20 & 11.20  & \cellcolor{gray!30}48.40\\
NIAH\_M2 &  19.40 &  33.40 &  18.40 & \cellcolor{gray!30}   67.40  & 17.20  & 25.00 & 0.6 & \cellcolor{gray!30}46.00& 4.60  & 1.60  &  0 & \cellcolor{gray!30}16.40\\
NIAH\_M3 &  9.80 & 14.00  &  1.20 & \cellcolor{gray!30}  45.80  & 12.60  & 3.20  & 0  & \cellcolor{gray!30}14.60&  4.00 & 0  & 0  & \cellcolor{gray!30}1.20\\
NIAH\_MQ &  93.55 & 75.30  &  84.70 & \cellcolor{gray!30} 97.85 & 95.30  & 59.40 & 0  & \cellcolor{gray!30}96.35&  48.85 & 6.05  &  0 & \cellcolor{gray!30}89.60\\
NIAH\_MV &  91.15 &  74.95 &  81.55 & \cellcolor{gray!30}  97.05   & 95.30  & 64.10 & 0  & \cellcolor{gray!30}96.60& 35.90  &  6.70 & 0.90  & \cellcolor{gray!30}89.90\\
VT       & 45.79  &  65.40 & 79.44  & \cellcolor{gray!30}  76.12   & 43.80  & 63.80 & 0 & \cellcolor{gray!30}79.24& 57.35  &  2.79 &  00 & \cellcolor{gray!30}62.04\\
CWE      &  34.78 & 51.59  & 7.56  & \cellcolor{gray!30}  34.89    & 0.26   & 21.11 & 0  & \cellcolor{gray!30}4.52& 0.02  &  1.20 &  0.02 & \cellcolor{gray!30}0.02\\
FWE      &  84.53 & 91.06  &  90.06 & \cellcolor{gray!30}  88.86   & 84.60  & 88.40 & 47.30 & \cellcolor{gray!30}87.53 & 85.00  &  0.73 & 56.53  & \cellcolor{gray!30}88.33\\
QA\_1    &  40.60 &  39.20 &  47.00 & \cellcolor{gray!30}   50.20   & 49.00   & 31.80 & 9.40 & \cellcolor{gray!30}55.00& 25.20  & 14.00  &  1.80 & \cellcolor{gray!30}57.40\\
QA\_2    & 33.60  &  33.00 &  34.20 & \cellcolor{gray!30}  38.20  & 34.00   & 30.40 & 8.00 & \cellcolor{gray!30}34.60& 22.20  &  16.60 &  0.60 & \cellcolor{gray!30}36.00\\
\bottomrule
AVg.     &  62.72 & 56.45 & 61.13   & \cellcolor{gray!30}\textbf{75.90}   & 61.95   & 47.01 & 5.02 & \cellcolor{gray!30}\textbf{69.46} & 35.97  &  15.96 &  12.17 & \cellcolor{gray!30}\textbf{59.48}\\
\bottomrule
  \end{tabular}
  }
   \caption{Comparison of per-task performance across different baselines on RULER at extrapolated lengths.}
   \label{tab:ruler_varies}
\end{table*}

\subsection*{E. Detailed per-task performance on RULER}
Table~\ref{tab:ruler_8K} presents RULER's per-task performance within the pretraining context window, and Table~\ref{tab:ruler_varies} shows the performance at extrapolated lengths.

\end{document}